\documentclass[sigconf]{acmart}
\usepackage{ctable}
\AtBeginDocument{%
  \providecommand\BibTeX{{%
    \normalfont B\kern-0.5em{\scshape i\kern-0.25em b}\kern-0.8em\TeX}}}

\usepackage{amsmath}
\usepackage{bbm}
\usepackage{enumitem}

\DeclareMathOperator*{\argmin}{arg\,min}

\graphicspath{{fig/}}
\setcopyright{rightsretained}
\copyrightyear{2020}
\acmYear{2020}
\acmDOI{}

\acmConference[KDD 2020 Workshop on MLG]{KDD 2020 Workshop on MLG}{August 22, 2020}{San Diego, CA, USA}
\acmBooktitle{16th International Workshop on Mining and Learning with Graphs, August 22, 2020, San Diego, CA, USA}
\acmPrice{}
\acmISBN{}

\begin{document}


\title{Hop Sampling: A Simple Regularized Graph Learning for Non-Stationary Environments}



\author{Young-Jin Park}
\email{young.j.park@navercorp.com}
\affiliation{%
  \institution{Naver R\&D Center, NAVER Corp.}
}

\author{Kyuyong Shin}
\email{ky.shin@navercorp.com}
\affiliation{%
  \institution{Clova AI Research, NAVER Corp.}
}

\author{Kyung-Min Kim}
\email{kyungmin.kim.ml@navercorp.com}
\affiliation{%
  \institution{Clova AI Research, NAVER Corp.}
}


\renewcommand{\shortauthors}{Young-J. Park, et al.}
\newcommand{\ky}[1]{\textcolor{blue}{#1}}
\newcommand{\km}{\textcolor{red}}

\begin{abstract}
Graph representation learning is gaining popularity in a wide range of applications, such as social networks analysis, computational biology, and recommender systems.
However, different with positive results from many academic studies, applying graph neural networks (GNNs) in a real-world application is still challenging due to non-stationary environments.
The underlying distribution of streaming data changes unexpectedly, resulting in different graph structures (a.k.a., concept drift).
Therefore, it is essential to devise a robust graph learning technique so that the model does not overfit to the training graphs.
In this work, we present \emph{Hop Sampling}, a straightforward regularization method that can effectively prevent GNNs from overfitting.
The hop sampling randomly selects the number of propagation steps rather than fixing it, and by doing so, it encourages the model to learn meaningful node representation for all intermediate propagation layers and to experience a variety of plausible graphs that are not in the training set.
Particularly, we describe the use case of our method in recommender systems, a representative example of the real-world non-stationary case.
We evaluated hop sampling on a large-scale real-world LINE dataset and conducted an online A/B/n test in LINE Coupon recommender systems of LINE Wallet Tab.
Experimental results demonstrate that the proposed scheme improves the prediction accuracy of GNNs.
We observed hop sampling provides 7.97 \% and 16.93 \% improvements for NDCG and MAP compared to non-regularized GNN models in our online service.
Furthermore, models using hop sampling alleviate the oversmoothing issue in GNNs enabling a deeper model as well as more diversified representation.
\end{abstract}

%
%


\keywords{Graph neural networks, Graph representation, Non-stationary graphs, Recommender System, Mobile Coupon Service}

\maketitle

\section{Introduction}

\begin{figure}[!t]
  \centering
  \vspace{0.5cm}
  \includegraphics[width=0.85\linewidth]{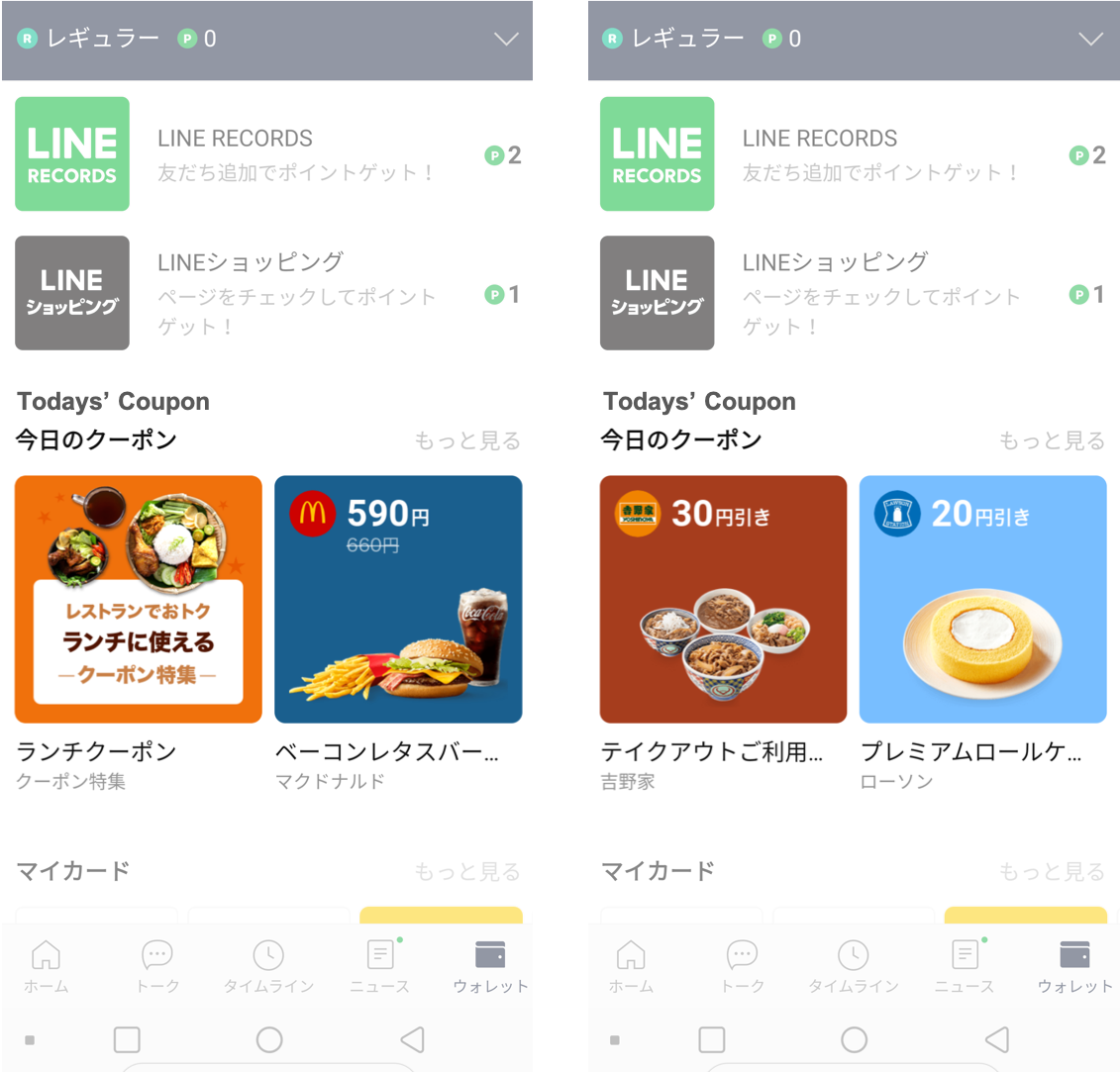}
  \caption{LINE Coupon Service. Everyday new coupons appear or disappear with sales events. User preference is heavily dependent on various factors such as discount ratio, and food type. We evaluate our graph learning method on such a non-stationary environment.}
  \label{fig:Line}
\end{figure}

Graph is an widely-applicable data structure, e.g. knowledge graphs, social networks, bioinformatics, recommender systems, etc. \citep{angles2008survey, hamilton2017representation}.
In recent years, graph neural networks (GNNs) \citep{scarselli2008graph} have recently been highlighted as a promising approach to handle graph-structured data, and have significantly improved diverse graph problems~\citep{grover2016node2vec, yu2017spatio}.
GNNs are broadly based on the message-passing algorithms, where each entity aggregates the representation vectors from its neighbors, recursively.
After $k$ aggregation steps, each node learns a new representation that is computed from the feature information of $k$-hop neighborhood as well as itself.
Consequently, GNNs are capable of capturing the structural information in the underlying graphs \cite{xu2018powerful}.

However, applying GNNs for real-world problems is not straightforward due to the following challenges.
Primarily, real-world environments are usually \emph{non-stationary} due to the evolving graph structures (a.k.a., concept drift).
Consider a recommender system, for instance, new users and items appear everyday, and the interaction between them varies as the users' item preferences change by external events \citep{xiang2010temporal} (see Table \ref{tab:stat}).
There often exists a time delay between the training and inference stages in many real-world services, the learning model would encounter unseen graphs in the online environment.
As such, we have to develop a flexible graph learning scheme that does not overfit to specific graphs that appeared during the training dataset.

Secondly, stacking deep GNN layers is non-trivial task.
In particular, Graph Convolutional Networks (GCNs) \cite{kipf2017semi}, the most prominent GNN model, are known to deliver the best performance when it is composed of two or one layers owing to the \emph{oversmoothing} problems \citep{kipf2017semi, klicpera2018predict}; it undermines the advantage of graph neural networks that can propagate node representations between arbitrary distant nodes. 
Such a problem becomes crucial in heterogeneous bipartite graphs that commonly emerges in real-world applications.
Taking the e-commerce as an example, the two-hop propagation in the user-item interaction graph is, in fact, equivalent to the one-hop graph among users.
Such shallow GNNs can not only lead to representation power degradation, but learn less distinct representation.

Recently, there have been several studies to address each of the presented challenges.
To resolve the first challenge, graph regularization techniques mostly based on the node sampling scheme has been researched \citep{hamilton2017inductive, huang2018adaptive, chen2018fastgcn, chiang2019cluster}.
On the other hand, to relieve the oversmoothing issue mentioned in the second challenge, variants of GNN with modifications on the propagation steps have been explored; APPNP \cite{klicpera2018predict} adopts the approximated personalized propagation scheme while JK-Nets \cite{xu2018representation} uses dense connections between GNN blocks of multiple hops.
Although suggested works could successfully resolve each challenge respectively in public datasets, we empirically found that those approachesLINE are not sufficient to fully handle non-stationary real-world datasets, as reported in Section \ref{sec:exp}.

In this paper, we propose \emph{Hop Sampling}, which is a simple but effective regularization scheme that can improve the previous GNNs to learn better graph representation that is applicable even in a non-stationary environment.
Hop sampling samples the number of aggregation/propagation steps during training stages rather than fixing it, unlike in most of the previous approaches.
Moreover, we would like to stress that the presented work is a complementary technique that can be applied together with the existing approaches.

We evaluate the hop sampling on a large-scale real-world dataset collected from LINE Coupon service (see~\figurename~\ref{fig:Line}), and run A/B/n test in our online services.
The experimental results demonstrate that the suggested model increases the ranking accuracy and allows the representation to be diversified, i.e., personalized.
We further found our model successfully avoid the oversmoothing issue even with the high number of propagation steps.

\begin{table}
  \caption{Statics of Line Coupon Dataset. First two columns in the table represents the percentage of intersection and new appearances of node/edge between the user-item interaction graph of two consecutive days, respectively. Third column describes the coefficient of variation regarding to the number of node/edge. As illustrated in the table, new nodes and edges appear in large proportions everyday and the number of nodes, and edges varies considerably, showing the serious non-stationarity of the dataset.}
  \label{tab:stat}
  \begin{tabular}{cccc}
        \toprule
                      & Intersection & New  & Coefficient of Variation \\ \hline
        Node          & 5.66 \% & 94.34 \% & 27.87 \% \\ \hline
        Edge          & 2.21 \% & 97.79 \% & 27.39 \% \\ 
        \bottomrule
    \end{tabular}
\end{table}


\section{Preliminaries}
\subsection{Graph Convolutaional Networks}

Consider a graph $G = (V, E)$ where $V$ and $E$ denotes nodes and edges, respectively.
The adjacency matrix is defined as $A \in \mathbb{R}^{|V|\times |V|}$ of which the element $A_{i,j}$ is associated to the edge $(v_i, v_j) \in V$.
To allow each node to gather its representation as well as its neighbors, the self loops are added to the adjacency matrix: \begin{math}\Tilde{A}=A+I\end{math}.

For given initial embedding matrix $X \in \mathbb{R}^{|V|\times D}$, graph convolutional network (GCN) updates the node representation by recursively aggregating embedding vectors of its neighbor:
representations are propagated as following:
\begin{align} \label{appnp_prop}
Z^{(k)} &= \sigma(\hat{A}Z^{(k-1)}W^{(k)}) \\
\hat{A} &= \Tilde{D}^{(-1/2)} \Tilde{A} \Tilde{D}^{(-1/2)}
\end{align}
where $Z^{(0)} = X$, $W^{(k)}$ is the learnable filter matrix of $k^{th}$ layer, $\sigma(\cdot)$ is the non-linear activation function (i.e., ReLU), and $\hat{A}$ is the symmetrically normalized adjacency matrix with self-loops with the diagonal degree matrix of $\Tilde{A}$, $\Tilde{D}$. 

\subsection{Approximate Personalized Propagation of Neural Predictions} \label{sec:appnp}


Despite the noticeable progress, previous GCN based approaches commonly focused on the shallow networks due to the oversmoothing.
Approximate personalized propagation of neural predictions (APPNP) \cite{klicpera2018predict} introduced a propagation scheme of personalized PageRank \cite{page1999pagerank} and relieved the issue.
APPNP add a teleport term to the root node in the graph propagation step, allowing the model to gather information from a far neighborhood while preserving the locality:
\begin{align}
Z^{(k)} &= (1-\alpha)\hat{A}Z^{(k-1)} + \alpha H \\ 
 &= \tilde{A}_k H = \Big( (1-\alpha)^k \hat{A}^k + \alpha \sum_{i=0}^{k-1} (1 - \alpha)^i \hat{A}^i \Big) H \label{eq:propagate}
\end{align}
where $Z^{(0)} = H = f_{p}(X)$ is the prediction matrix calculated by the neural network $f_{p}$ and $\alpha$ is the teleport probability.
Note that, APPNP separates the prediction and propagation stage so that the model does not have learnable parameters during the propagation stage, which helps the model to avoid overfitting.



\section{Proposed Method: Hop-Sampling}


Previous studies, including GCN and APPNP, have focused on the dataset of which the graph is fixed between train and test set.
As such, most GNN approaches are inherently transductive; they often fail to generalize for unseen graphs \cite{hamilton2017inductive}.
In the real-world applications, however, the environment is often non-stationary, therefore a robust \emph{inductive} graph learning scheme that does not overfit to the train graphs is required.
To tackle the issue, we propose a simple but effective regularization method, \emph{Hop Sampling}:
In the hop sampling scheme, we randomly select the number of propagation steps from $0$ to $K$ for every batch instead of fixing it.
Consequently, the proposed scheme optimizes the expectation of embedding instead of the final embedding:
\begin{equation}
\label{hop_samp}
\theta^* = \argmin_{\theta} \sum_{t \in \mathcal{T}} l_\theta(Z = \mathbb{E}_{k \sim p_{samp}(\cdot)}\big[ Z_t^{(k)}\big])
\end{equation}
where $l_\theta$ is a loss function of learnable paramater $\theta$ in GNNs and embedding networks, and $p_{samp}$ is a sampling distribution.
Note that our model is equivalent to the existing approaches when the sampling distribution is an indicator function: $\mathbbm{1}(k=K)$.
In this paper, we adopted a discrete uniform distribution $\mathcal{U}(0, K)$ instead.
Future work includes considering non-uniform parameterized sampling probability distribution (i.e., learn to select the hop).
When validation and testing, we do not sample $k$.

\textbf{Regularization Perspectives.} \hspace{0.2cm}
It is recently studied that the graph representations tend to crumble after multiple propagation steps, especially in GCN based approaches. 
We believe one of main factors in the problem is that existing approaches lacks the signal helping the model to learn meaningful representation on intermediate propagation stages.
In the meantime, hop sampling prevents GNNs from overfitting to a specific hop number and allows the model to learn informative embeddings at every propagation step.  

\textbf{Graph Sampling Perspectives.} \hspace{0.2cm}
Remark that in the propagation scheme, we can consider $k$ as a factor transforming the adjacency matrix propagating the initial feature matrix $H$, from $\hat{A}$ to $\tilde{A}_k$ as illustrated in eq \eqref{eq:propagate}.
In that sense, hop sampling can be regarded as a new graph sampling technique when it is applied to APPNP.
Consequently, the model using hop sampling experiences $K$ times larger variety of graphs during the training stage, helping the model to adapt a non-stationary environment easily and avoid overfitting.


\section{Experiment Results} \label{sec:exp}
\begin{table}
  \caption{Hyperparameters.}
  \label{tab:hyper}
  \begin{tabular}{cc}
    \toprule
    Hyperparameters&Values\\
    \midrule
    Teleport probability $(\alpha)$ &  0.3\\
    Dimension of node embedding $z_i$ ($D$) & 128\\
    \# of propagation $(K)$ &  [1, 2, 4, 8, 16] \\
    Batch size & 1024 \\
    Initial learning rate & $3e^{-6}$\\
    $\beta_1$ of Adam & 0.9\\
    $\beta_2$ of Adam & 0.999\\
  \bottomrule
\end{tabular}
\end{table}

To demonstrate the effectiveness of hop sampling in real-world non-stationary applications, this paper addresses the use case of recommender systems as one of the representative examples.
Utilizing GNNs in recommender systems has been recently received attention because of its capability to model the relational structure between user and item \citep{berg2017graph, ying2018graph, wang2019knowledge, kim2019tripartite, shin2020multi}.
Consider a recommender system with $N_u$ users and $N_i$ items. 
Each user and item becomes a node in bipartite graph $G_t$, of which the nodes are connected if there is positive interaction (e.g., click or use) between the corresponding user and item during a period of time $t \in \mathcal{T} = \{T_1, \cdots, T_L\}$.
In most real-world cases, the graph $G_t$'s are heterogeneous because users and items have different types of side-information.
For example, user nodes may contain demographic properties such as age and gender, while item nodes may have item categories and visual-linguistic contents.
Denote the side information of users $Y_u \in \mathbb{R}^{N_u\times M_u}$ and side information of items $Y_i \in \mathbb{R}^{N_i\times M_i}$. \footnote{In this paper, we assumed the side information is invariant during time. Without loss of generality, this can be extended to time-variant data.}
Then, the node feature matrix $X$ of each node type is obtained by separated embedding networks $f_u$ and $f_i$:
\begin{equation} \label{eq:feature}
X = \left[ f_u(Y_u) ; f_i(Y_i) \right] \in \mathbb{R}^{(N_u + N_i) \times D} .
\end{equation}

Starting from the node feature, GNN transforms the node representation into $Z_t$ through $K$ propagation steps using the graph $G_t$.
The role of GNN is to improve the node representation of each root node by gathering information from its neighborhood and features of itself.
As a consequence, each node embeddings in bipartite graph are mixed by neighborhood items and users.
Finally, we predict the preference scores between user and item after $t$ by computing the inner-product between embeddings of them:
\begin{equation} \label{eq:predict}
p_{t, i, j} = sigmoid(z^T_{t, i} z_{t, j}),
\end{equation}
where \begin{math}z_{t, i}\end{math} is \textit{i}-th row vector of the final node representation matrix \begin{math}Z_t\end{math}.
The model parameters are optimized to minimize the cross-entropy loss between predicted score and true interaction label for all time periods via stochastic gradient descent algorithms.

 \label{sec:user}

\subsection{Dataset}
We use a dataset collected from a large-scale coupon recommendation system in LINE service as described. 
We constructed the bipartite graph consists of those 120,000 user and 517 item nodes connected when the user used the corresponding coupon.
\footnote{Since the number of total users using the service is huge (over 10 million), we collected a subset of users as a user pool. We empirically found that using a user pool not only reduces the memory requirements but helps faster convergence of the model.}
As side information, we used gender, age, mobile OS type, and interest information for users, while brand, discount information, test, and image features for items.
For each attribute, the shallow multi layer perceptrons (MLPs) are used to make $n$-dimensional feature vector, and we aggregate them with sum operation to obtain feature matrix $X$.
We split a dataset on daily basis: the first 14 days, the subsequent 3 days, and the last 3 days as a train, valid, and test set, respectively, i.e., ${\scriptstyle |\mathcal{T}_{train}|=14, ~~ |\mathcal{T}_{valid}|=3, ~~ |\mathcal{T}_{test}|=3}$.
For each day, we construct bipartite graphs $G_t$ by using interactions in the previous 28 days.
As expected, interactions in valid and test set are masked on the graph during the experiments.

\subsection{Comparable models}
To demonstrate the recommendation performance of the proposed method, we compared our model with following models:
\begin{itemize}[leftmargin=*]
    \item \textbf{Deep Neural Networks (DNNs)} 
    is equivalent to 0-hop GNN model, i.e., $K$=0, and compared to show the effect of the graph propagation on the recommender system.

    \item \textbf{GCN}~\cite{kipf2017semi}
    GCN is one of the most widely used graph neural networks in the literature.
    GCN filters out noises in the graph based on graph signal processing.
    
    \item \textbf{Jumping Knowledge Networks (JK-GCN)} \cite{xu2018representation}
    concatenates GCN blocks of multiple hops. This method helps to alleviate oversmoothing with skip connections by adaptively adjust aggregation range of neighbor nodes. We denote this model as JK-GCN in this paper.
    
    \item \textbf{APPNP}~\cite{klicpera2018predict} adopts propagation scheme described in Section \ref{sec:appnp}. 
    We fix the prediction function $f_\theta$ as identity function, since we already have learnable parameters in $f_u$ and $f_i$. 
    
    \item \textbf{APPNP-Hop Sampling (HS)}
    is the proposed model which applies propagation scheme and hop sampling together.
    To further show the generalizability, we also report the performance of GCN model with hop sampling, \textbf{GCN-HS}.
    
\end{itemize}
We applied node sampling techniques for every GNN model for the scalable and inductive graph learning; we sampled 10240 users uniformly at random \footnote{Empirically found that degree sampling shows the same result as uniform in our cases} for each batch.
For the fairness of the comparison, we adopted the same network architectures for $f_u$ and $f_i$.
Important hyperparameters are summarized in Table \ref{tab:hyper}.
Experiments were performed on NAVER SMART Machine Learning platform (NSML) \cite{kim2018nsml, sung2017nsml} using PyTorch \cite{NEURIPS2019_9015}.

\setlength{\tabcolsep}{9pt}
\ctable[
    caption = {Average values for ranking metrics of 20 runs on the LINE dataset. 
},
    label = tab:performance,
 	doinside=\normalsize
]{lcccc}{
}{
\toprule
Models & NDCG@10 & MAP@10  & HIT@10  \\
\midrule
DNN & 0.1685  & 0.1337  & 0.3227    \\
GCN & 0.2114  & 0.1652  & 0.4214    \\
JK-GCN & 0.1583 & 0.1208 & 0.326 \\
APPNP & 0.2919 & 0.2344  & 0.5852  \\
\midrule
GCN-HS & 0.2124 & 0.1685 & 0.4307 \\
APPNP-HS & \textbf{0.3039}  & \textbf{0.2487} & \textbf{0.5922}    \\
\bottomrule
}


\setlength{\tabcolsep}{9pt}
\ctable[
    caption = {Average values for diversity metrics of 20 runs on the LINE dataset. 
},
    label = tab:diversity,
      doinside=\small
]{lcccc}{
}{
\toprule
Models  & ILD@10 & Coverage@10 & Entropy@10   \\
\midrule
DNN & 0.1441 & 0.0285 & 3.622    \\
GCN & 0.446 & 0.2923 & 4.733    \\
JK-GCN & 0.6288 & 0.4846 & 5.519 \\
APPNP & 0.7866 & 0.8434 & 6.629  \\
\midrule
GCN-HS & \textbf{0.8536} & \bf 0.9331 & \textbf{7.151} \\
APPNP-HS & 0.8229 & 0.8494 & 6.845    \\
\bottomrule
}


\setlength{\tabcolsep}{9pt}
\ctable[
    caption = {Relative online performance improvement to Top Popular recommendation for ranking metrics in LINE Coupon recommender systems. 
},
    label = tab:online,
 	doinside=\normalsize
]{lcccc}{
}{
\toprule
Models & NDCG@10  & MAP@10  & HIT@10              \\
\midrule

APPNP & 76.90 \% & 82.40 \% & 61.24 \% \\
APPNP-HS & \bf 90.99 \% & \bf 113.28 \% & \bf 65.05 \% \\
\bottomrule
}

\subsection{Experimental Results and Analysis}
Primarily, we report three ranking metrics, Normalized Discounted Cumulative Gain (NDCG) \cite{jarvelin2002cumulated}, mean Average Precision (MAP), and  Hit Ratio (HIT) to evaluate the ranking accuracy of the models.
Since the optimal value for the number of propagation $K$ differs by the model, we chose models with the highest NDCG@10 value; 
the optimal $K$'s were 2 for GCN and JK-GCN, 4 for GCN-HS and APPNP-HS, and 8 for APPNP.
As shown in Table \ref{tab:performance}, our models further enhances the prediction power for every ranking metric compared to GNN models only with node sampling.

Furthermore, we measured three diversity metrics, Inter-List Diversity (ILD) \cite{zhou2010solving}, item coverage, and Shannon entropy for each model to show how diversified the recommended items are.
As shown in Table \ref{tab:diversity}, our model shows the highest diversity values for every metric, which indicates that hop sampling relieves oversmoothing and encourages the model to learn graph representation in a more diversified way.

\begin{figure}[!t]
  \centering
  \includegraphics[width=\linewidth]{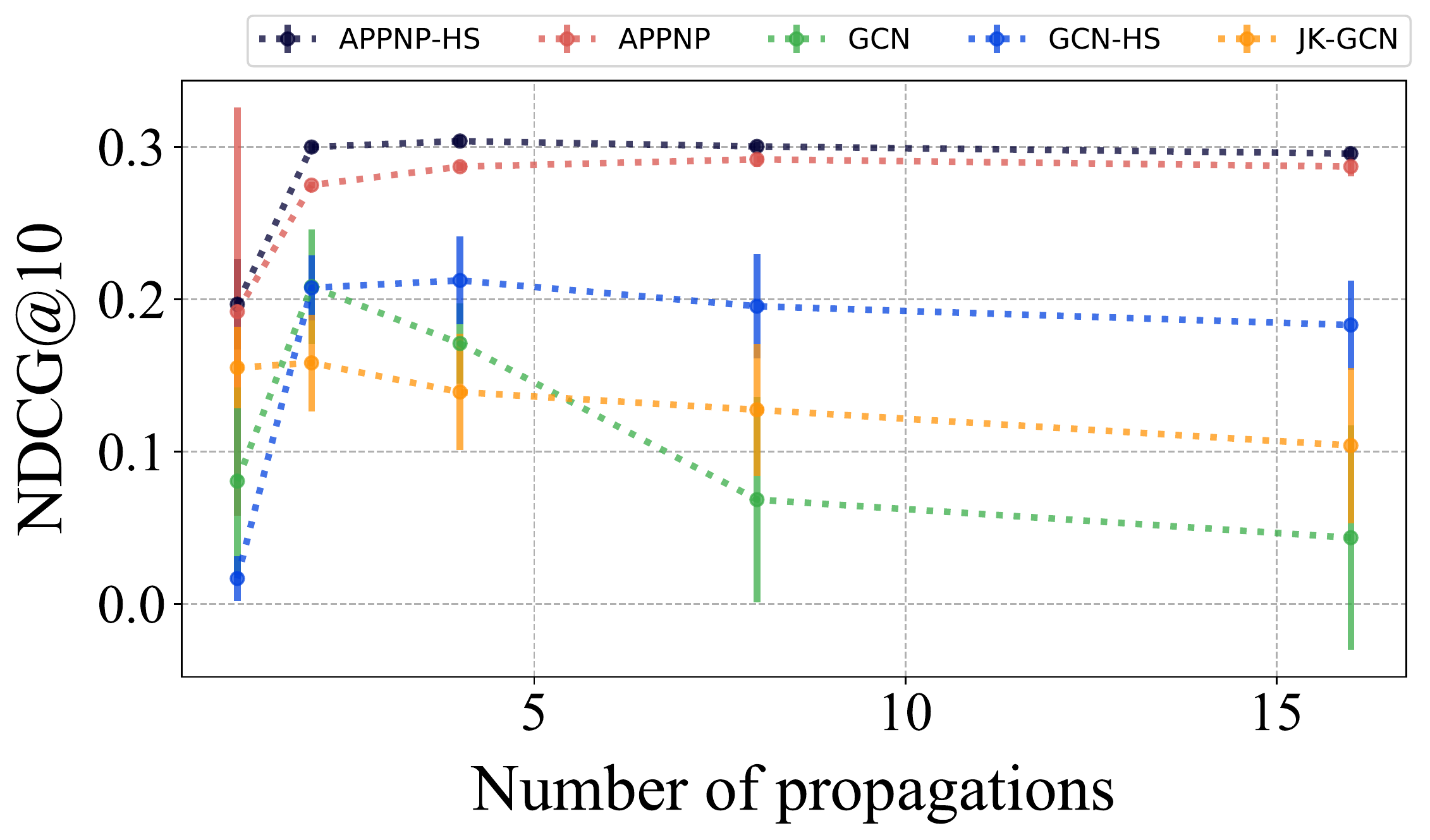}
  \caption{NDCG@10 by the number of propagation steps on a LINE dataset. Error bar indicates 95\% confidence interval.}
  \label{fig:performance}
\end{figure}

\begin{figure}[!t]
  \centering
  \includegraphics[width=\linewidth]{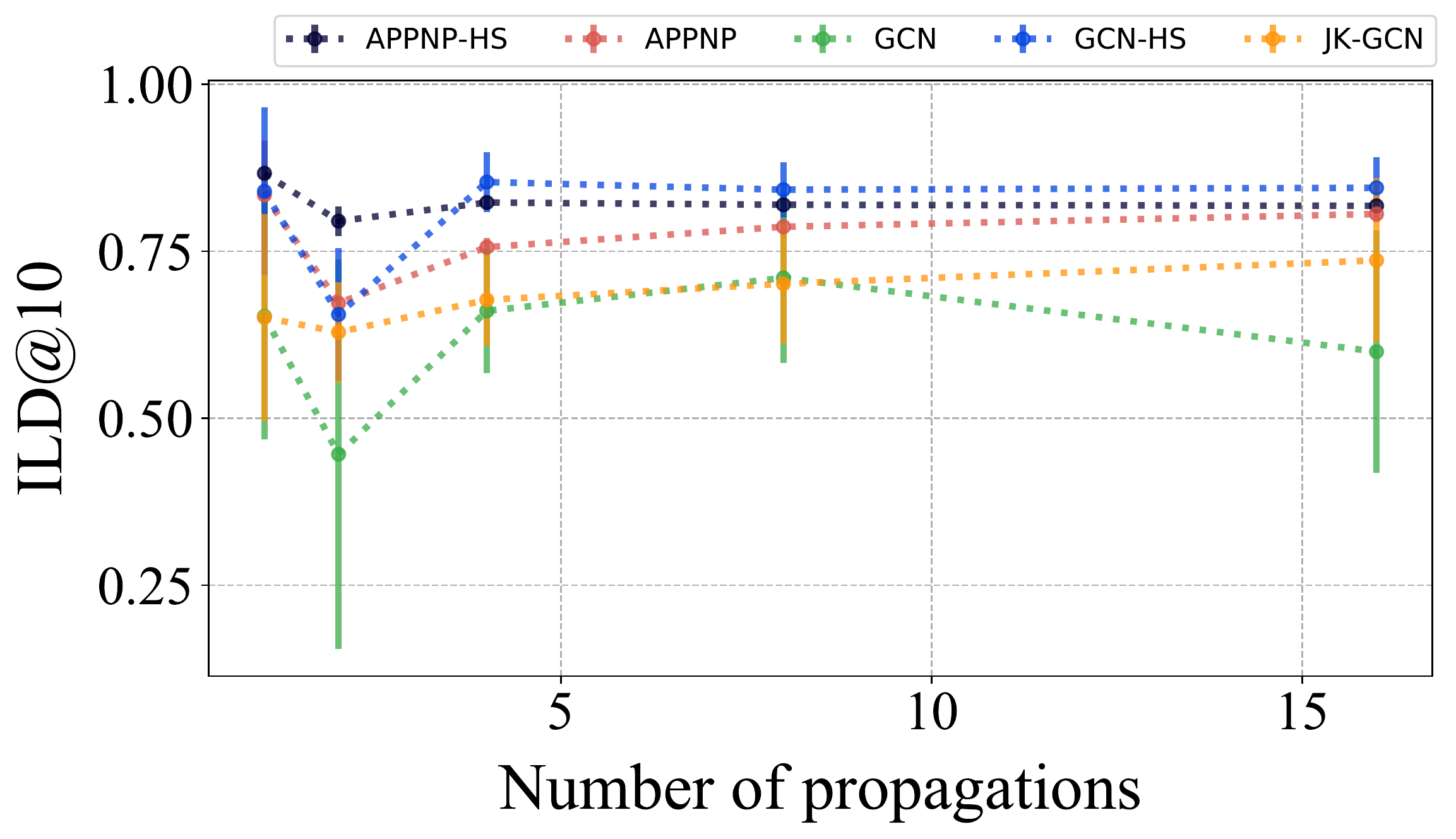}
  \caption{ILD@10 the number of propagation steps on a LINE dataset. Error bar indicates 95\% confidence interval.}
  \label{fig:diversity}
\end{figure}

The prediction performance and recommendation diversity with different propagation steps are presented as well.
Figure \ref{fig:performance} and Figure \ref{fig:diversity} show that hop sampling enables the model to stack deeper layers for both GCN and APPNP whereas GCN and JK-GCN fail to learn graph representation over two hops as expected.
As a result, both the prediction power and the diversity of hop sampling models are enhanced.
Moreover, we found that hop sampling reduces variance of GNNs.
Overall, APPNP-HS, with $K$=4 shows the best graph representation in our experiment.

\begin{figure}[!t]
  \centering
  \includegraphics[width=\linewidth]{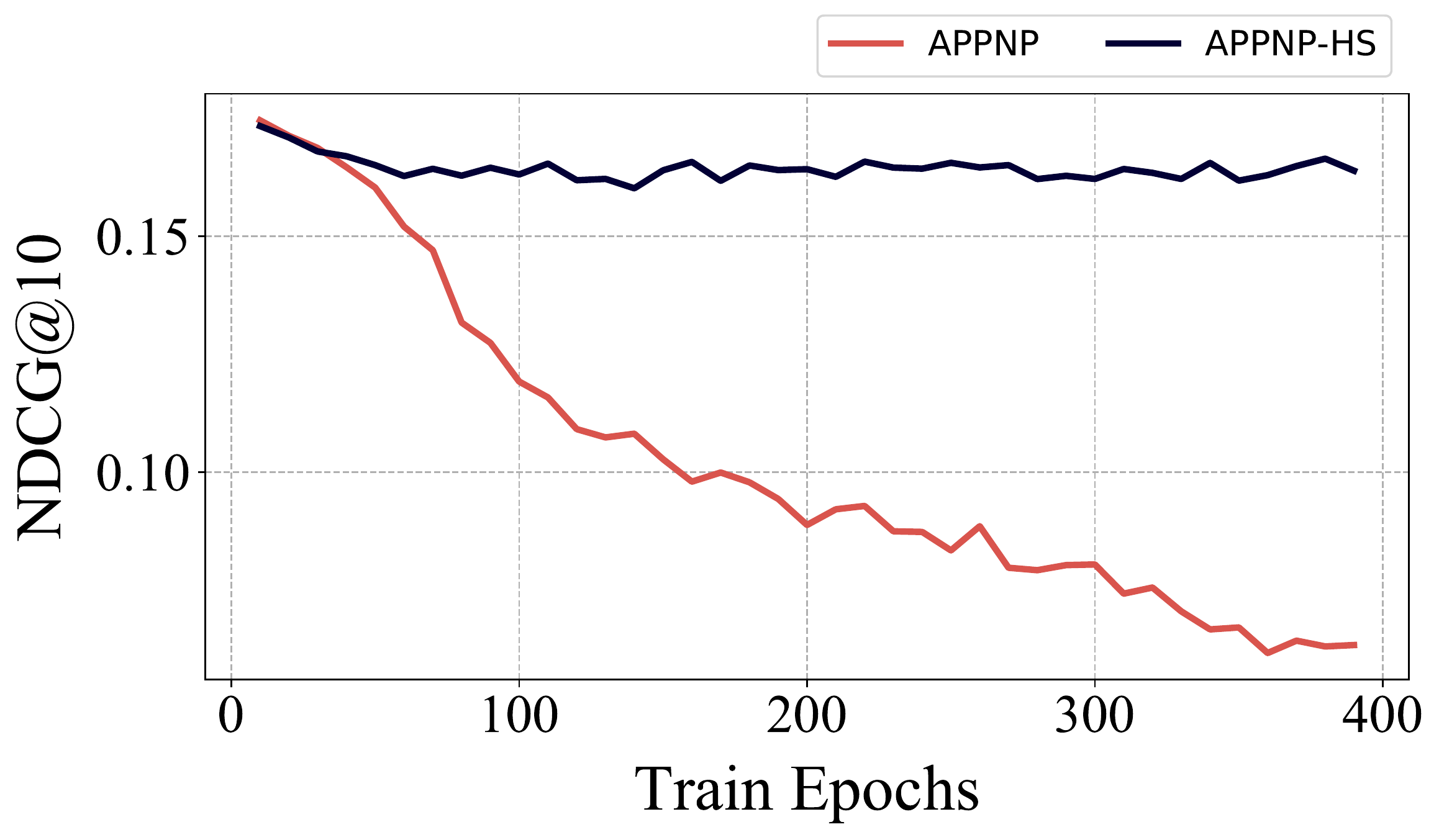}
  \caption{NDCG@10 by the train epochs on a LINE dataset with empty test graphs.}
  \label{fig:attack}
\end{figure}

Finally, we conducted a synthetic experiment to see the effect of hop-sampling in extremely non-stationary conditions.
In real-world cases, we have experienced the situation that the empty user-interaction graph is given to our systems due to some practical issues such as log data omission.
Similarly, we hypothetically remove edges of graphs in test datasets and compared the result between APPNP and APPNP with hop sampling.
Figure \ref{fig:attack} shows that APPNP without hop sampling suffer from the overfitting while the test accuracy continuously decreases over the train epochs.
We believe the hop sampling helps the model not to rely on the training graph overly, and the initial embedding matrix $X$ is learned more robustly and adequately.

We have deployed best performing GNN models on offline test, APPNP and APPNP-HS, in LINE Coupon service and run A/B/n tests to evaluate the online performance on our LINE Coupon recommender systems in LINE Wallet Tab for four days.
Relative performance improvement to non-personalized Top Popular recommendation is reported in Table \ref{tab:online}.
As shown in results, GNNs provides significant improvement to our systems, and hop sampling further enhances the recommendation performance by 7.97 \%, 16.93 \%, and 2.37 \% for NDCG, MAP, and HIT, respectively, compared to the APPNP with node sampling.

\section{CONCLUSION}
In this paper, we presented Hop Sampling, a novel and straightforward technique helping models to learn better graph representation.
By varying the number of propagation steps randomly, hop sampling alleviates the overfitting and oversmoothing problems in GNNs.
Experimental studies on real-world, large-scale LINE Coupon recommender system shows the proposed scheme improves the recommendation quality in terms of both ranking accuracy and recommendation diversity.


\begin{acks}
The authors appreciate to NAVER Clova ML X team for insightful comments and discussion.
\end{acks}

\bibliographystyle{ACM-Reference-Format}
\bibliography{references}


\end{document}